\pdfoutput=1

\documentclass[11pt]{article}

\usepackage{ACL2023}

\usepackage{times}
\usepackage{latexsym}

\usepackage[T1]{fontenc}

\usepackage[utf8]{inputenc}

\usepackage{microtype}

\usepackage{inconsolata}

\usepackage{booktabs}

\usepackage{graphicx}

\usepackage{amsmath}

\begin{document}

\title{In-Context Learning User Simulators for Task-Oriented Dialog Systems}


\author{Silvia Terragni, Modestas Filipavicius, Nghia Khau, \\ \textbf{Bruna Guedes, André Manso \and Roland Mathis}\\
Telepathy Labs GmbH\\ Zürich, Switzerland\\
\texttt{\{firstname.lastname\}@telepathy.ai}
}

\maketitle

\begin{abstract}
This paper presents a novel application of large language models in user simulation for task-oriented dialog systems, specifically focusing on an in-context learning approach. By harnessing the power of these models, the proposed approach generates diverse utterances based on user goals and limited dialog examples. Unlike traditional simulators, this method eliminates the need for labor-intensive rule definition or extensive annotated data, making it more efficient and accessible. Additionally, an error analysis of the interaction between the user simulator and dialog system uncovers common mistakes, providing valuable insights into areas that require improvement. 
Our implementation is available at \url{https://github.com/telepathylabsai/prompt-based-user-simulator}.


\end{abstract}

\section{Introduction}
Task-oriented dialog systems~\cite[TODS]{zhang2020survey} are designed to assist users in accomplishing tasks, such as booking a flight or ordering food. These systems need to be able to understand user intentions and generate appropriate responses. 
User Simulators (US) are essential components of TODS, as they enable researchers and developers to test and evaluate the system's performance in a controlled environment before deploying it to real users. Yet, building a simulator can be a time-consuming and costly process, as it requires the definition of rules to generate user actions~\cite{schatzmann2007agenda,Eckert1997bigramUS}, or a large amount of annotated data to train or fine-tune a model~\cite{Lin2022GenTUSSU,Asri2016seq2seqUS}. 

Large Language Models~\cite[LLM]{Liu2021PretrainPrompt}, such as GPT-4 \cite{openai2023gpt4}, exhibit few-shot or in-context learning capabilities. They need just a few \textit{context} examples to perform a new task, without any fine-tuning. In-context learning has also been applied to a wide variety of tasks, including dialog system tasks, such as intent classification, dialog state tracking, and natural language generation~\cite{madotto2021few,Li2022DialogSimulation,Madotto2020LanguageMA,Dingliwal2021FewSD}. These advancements open up new opportunities for leveraging LLM capabilities in user simulators for TODS, promising more efficient development and evaluation processes.

In this paper, we propose a novel application of LLMs for user simulation in dialog systems. Our approach leverages the few-shot learning capabilities of LLMs, where the user simulator requires a user goal, representing the objectives the user intends to achieve, and a small set of dialog examples. By utilizing this approach, the model can generate diverse and contextually appropriate utterances in natural language. This capability enables the simulator to explore different scenarios and tackle diverse and unexpected user inputs.

A notable advantage of our proposed approach is that it does not require fine-tuning, making it transferable across multiple domains. We evaluate the effectiveness of our approach using the MultiWOZ dataset, which encompasses dialogs from various domains. Furthermore, we conduct an error analysis to identify common mistakes that arise from the interaction between a prompt-based user simulator and a dialog system. This analysis helps uncover key areas for improvement in the development of TODS.

\section{Related Work} \label{soa}

\subsection{User Simulation Approaches}

One of the earliest US approaches is the Bigram model, which computes the probability of user actions given system actions~\cite{Eckert1997bigramUS}. The model assumes that user actions are solely dependent on system actions and does not account for the coherence of user goals. 
To address this issue, rule-based methods have been developed, which rely on rules based on heuristics or expert knowledge. Agenda-based simulation is an example that uses a stack of actions to simulate the user's goals~\cite{schatzmann2007agenda,Schatzmann2009agenda,Keizer2010agenda}. These models can generate realistic user behavior, but they require manual rule definition and are not transferable to new domains.

Data-driven approaches have gained popularity due to the availability of large-scale annotated datasets and the development of advanced machine learning techniques. These approaches use deep learning models, such as recurrent neural networks~\cite{Gur2018rnnUS,Asri2016seq2seqUS} or transformers~\cite{Lin2021DomainindependentUS,Lin2022GenTUSSU}, to learn from data and generate user actions~\cite{Lin2021DomainindependentUS,Asri2016seq2seqUS} or utterances in natural language~\cite{Lin2022GenTUSSU,Kreyssig2018seq2seqUS,Cheng2022finetuneUS,Liu2022GPTusersim}. In addition to these approaches, a parallel line of research explores the use of user simulators into conversational information retrieval systems~\cite{Sekulic2022EvaluatingMC,Wang2023AnII}. 

Data-driven models do not rely on predefined rules and can capture complex patterns, such as the coherence user goals and the diversity of user responses. However, they require significant computational resources to train or fine-tune the models and require annotated data, which can be expensive and time-consuming to create.

\subsection{Large Language Models and Prompting}
The recent advancements in Large Language Models (LLMs) have been remarkable~\cite{Bowman2023EightThingsLLM}, with OpenAI's GPT-3~\cite{Brown2020GPT3} leading the way as the first modern general-purpose LLM. It exhibits few-shot (or in-context) learning abilities, adapting to new tasks with few examples. Prompting in LLMs~\cite{Liu2021PretrainPrompt} is providing a query to guide the model's text generation. It can be a question, statement, or instruction that sets the context for the model's response. The prompt influences the generated text's content, style, and tone. Manual creation of prompts based on human introspection is commonly used~\cite{Schick2020ExploitingCF,Schick2020ItsNJ}. 


The advent of ChatGPT~\cite{openai2023chatgpt} has revolutionized the field of Conversational AI and every week GPT-like models are released~\cite{alpaca2023,huggingchat2023,openassistant2023}, significantly increasing the interest in prompt-based and in-context learning approaches. While previous works have explored the use of LLMs for dialog generation~\cite{Li2022DialogSimulation} or dialog systems~\cite{madotto2021few,Shin2022DialogueSummarization}, to the best of our knowledge, there has been limited exploration of the few-shot capabilities of these models for user simulation~\cite{Li2022DialogSimulation}. The ability of LLMs to generate human-like responses and adapt to new tasks with minimal examples provides exciting opportunities for creating more realistic and dynamic conversational experiences. 


\section{Background}

\subsection{Task-Oriented Dialog Systems}
TODS usually rely on an ontology that includes all possible \textit{intents, domains, slots}, and \textit{values} that users or the system can use in their actions. Domains describe the entities the user or system talk about and are characterized by slots that can have multiple values. Users and systems may have different intents, such as users informing the system about their needs or systems providing recommendations. These intents can be expressed in natural language (e.g., \textit{``I am looking for an Italian restaurant with a moderate price range''}) and translated into lists of semantic actions, also referred to as user and system acts: 
\begin{align*}
    [&(Inform, Restaurant, Food, Italian), \\&(Inform, Restaurant, Price, moderate)]
\end{align*}

\subsection{User Goal}
User goal is fundamental for TODS as it refers to the objectives that a user wants to achieve through their interaction with the system. A user goal can be represented as a list of \textit{(intent, domain, slot, value)} tuples. 
For example, in a restaurant reservation system, a user goal could be to make a reservation for a specific date for two people in an Italian restaurant. The resulting user goal can be represented as 
\begin{align*}
    [&(Inform, Restaurant, Food, Italian),\\&(Book, Restaurant, People, 2),\\&(Book, Restaurant, Date, Monday)].
\end{align*}

A US should fulfill all the user goal requirements by the end of the dialog. We can easily convert the structured user goal representation into natural language using templates, which are predefined patterns that can be filled with the relevant structured information. This natural language representation of user goals becomes particularly relevant when leveraging LLMs for user simulation.

\begin{table*}[h]
\centering
\caption{One-shot prompt examples for generating three turns of a dialog. The sentences in \textit{italic} are generated by a user simulator, then they will be part of the prompt for the next turn.}
\resizebox{1\textwidth}{!}{%
\begin{tabular}{ll}
\toprule
Initial prompt & \begin{tabular}[c]{@{}p{15.5cm}@{}}These are examples of conversations between a CUSTOMER and an ASSISTANT. Complete the conversation as a CUSTOMER. The conversation is about making a reservation or asking for information according to the REQUIREMENTS. You must fulfill all REQUIREMENTS in the given order. Be precise with the REQUIREMENTS, clear and concise. \\ \\ Example 1:        \\ REQUIREMENTS: You are looking for a train departing from Peterborough and arriving in Cambridge by 19:30 on Sunday. \\ CUSTOMER: I 'm looking for a train that departs from peterborough and arrives by 19:30    \\ ASSISTANT: Where would you like to go to ?     \\ CUSTOMER: I am going to Cambridge on Sunday .    \\ ASSISTANT: There's a train that arrives at 19:09. Would that do?                      \\ CUSTOMER: Sure would. What time does it depart and how much does it cost?           \\ ASSISTANT: leaves at 5:19 13.2 GBP payable at the station your reference number is CLPW6OR9
\\ CUSTOMER: thanks for the service, that is all I need.\\ \\ Example 2:\\ REQUIREMENTS: You are looking for a restaurant that serves British food and is in the South. Make sure you get the phone number and postcode.\\ CUSTOMER:\end{tabular} \\
\midrule
Prompt at turn 1 & \begin{tabular}[c]{@{}p{15.5cm}@{}}\textit{Hi, I'm looking for restaurants in the area that serve British cuisine.}\\ ASSISTANT: There are many British restaurants in the city. Could you tell me in what area you would like to eat?\\ CUSTOMER:\end{tabular} \\
\midrule
Prompt at turn 2 & \begin{tabular}[c]{@{}p{15.5cm}@{}}\textit{In the south}\\ ASSISTANT: We have 7 such places. Restaurant one seven has some great reviews. The phone number is 01223337766.\\ CUSTOMER:
\end{tabular}
\\\bottomrule
\end{tabular}%
}
\label{tab:example_prompt}
\end{table*}

\section{Proposed Method}
Hereby, we present an approach for an in-context learning user simulator. The simulator generates user utterances based on example dialogs, the target user goal, and the conversation history. We also investigate random and similarity-based strategies for selecting example dialogs. 

\begin{figure*}[h]
    \centering
\includegraphics[width=0.6\linewidth]{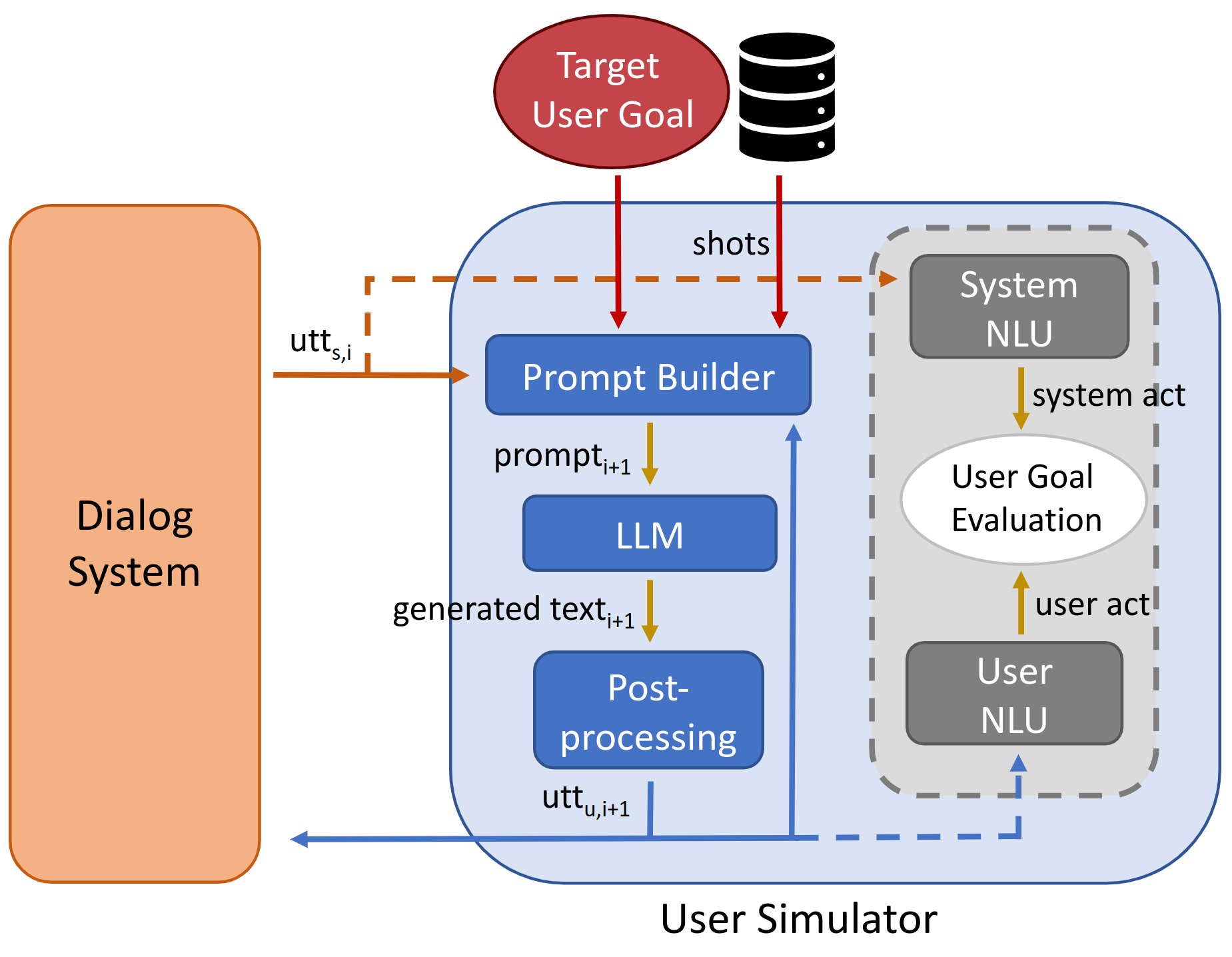}
    \caption{System and user simulator architecture sketch.}
    \label{fig:architecture}
\end{figure*}

\subsection{A Prompt-based User Simulator}
The aim of the US is to generate user's utterances given a target user goal $g_t$. The simulator interacts directly with the dialog system, thus requiring the US to generate an utterance per turn. 
Consider a target dialog $d_t$, i.e. the conversation to be generated by the interaction between US and system. It is defined as a sequence of interactions as follows:
\begin{equation}
    [(utt_{u,1}, utt_{s,1}), \ldots, (utt_{u,N}, utt_{s,N})]
\end{equation}
 where $utt_{\cdot, i}$ denotes an utterance in natural language from user $u$ or system $s$ at turn $i \in 1, \ldots, N$ (where $N$ is the number of turns). The generation of the user utterance $utt_{u,i}$ at turn~$i$ of the target dialog $d_t$ is conditioned on: 
\begin{itemize}
    \item $K$ shots, which are pairs of user goal and the corresponding dialog, $(g_1, d_1),\ldots,(g_K, d_K)$; 
    \item the target goal $g_t$ of the dialog $d_t$ to generate;
    \item the target dialog $d_t$ until turn $i_{i-1}$, i.e. $[(utt_{u,1}, utt_{s,1}), \ldots, (utt_{u,i-1}, utt_{s,i-1})]$.
\end{itemize}
Table~\ref{tab:example_prompt} shows an example of the prompts used by the LLM to generate the user texts. In the first row, we can see the initial prompt, starting with a task description. Next, we add an example dialog with the corresponding user goal and the target goal $g_t$. The goal $g_t$ can be defined by a domain expert or generated based on system requirements. The prompt ends with the word \textit{``CUSTOMER:''} to enforce the LLM to generate a user utterance. 

In the first turn of the conversation, the US generates an utterance, conditioned on the prompt. This utterance serves as input for the dialog system. The two utterances are then appended to the initial prompt, thus composing the \textit{prompt at turn 1}. Next, US generates a new utterance based on the updated prompt and the interaction continues until the conversation ends. 




\subsection{Shot Selection}\label{sec:shot_sampling} 
Given the user goal~$g_t$ of the dialog to generate, we investigate two strategies from sampling the dialog examples that would serve as shots, following~\cite{Li2022DialogSimulation}. 
The first strategy consists in random sampling. As second strategy, we sample the examples based on a similarity criterion on the user goal. In particular, we compute the Jaccard similarity between the domain and the slot sets of the target user goal~$g_t$ and a sampled user goal~$g_k$. 
More formally, the Jaccard similarity $J({g_k},{g_t})$ between the $g_k$ and $g_t$ can be defined as follows:
\begin{equation}\label{eq:dialog_similarity_score}
J({g_k},{g_t}) = \dfrac{|Dom_{g_t} \cap Dom_{g_k}|}{|Dom_{g_t} \cup Dom_{g_k}|}\cdot\dfrac{|Slot_{g_t} \cap Slot_{g_k}|}{|Slot_{g_t} \cup Slot_{g_k}|}
\end{equation}
where $Dom_{g_*}$ and $Slot_{g_*}$ are respectively the set of domains and slots of the goal of the dialog $*$, and the $|S|$ denotes the cardinality of a set $S$. We compute the similarity between the target goal and the sampled goals and select the $K$ most similar goals.

\subsection{Overview of the Architecture} \label{sec:architecture}

Figure~\ref{fig:architecture} shows the architecture for the interaction between the user simulator and the system. The dialog system can be thought of as a black-box that returns a natural language utterance per turn. On the US side, the prompt builder takes as input the dialog history (i.e. system and user utterances at turn $i$ ($utt_{s,i}, utt_{u,i}$)), the target user goal $g_t$, and the shots. The prompt builder then generates the prompt for turn $i+1$. The LLM generates a text based on the prompt, which is post-processed to ensure that it only contains user utterances. The resulting text is returned to the dialog system.

The architecture also includes an evaluation component (the grey box), which is not strictly necessary for the US to generate new utterances and complete a conversation. It tracks the goal fulfillment and evaluate the system's interaction. This component consists of system and user-side NLUs for translating the system and user utterances, respectively, into dialog acts (DAs). These DAs are then used to assess the user goal fulfillment.

\section{Experimental Setting}
We carry out our experiments using ConvLab-2 \cite{zhu2020convlab2}. The framework provides models and evaluation tools for developing a simulator\footnote{Our implementation is available at \url{https://github.com/telepathylabsai/prompt-based-user-simulator}.}. It also includes a generator for the target user goals. 

\subsection{Dataset}

We evaluate the proposed models on the well-known MultiWOZ 2.1 dataset~\cite{eric2019multiwoz}, available in ConvLab-2. The dataset contains 10,433 dialogs between an agent and a user, and each dialog is about one or more of 7 domains. 
The dataset contains system and user acts, represented as $(domain, intent, slot, value)$. 
Dialogs are characterized by user goals in natural language. 

\subsection{Goal Fulfillment Evaluation Metrics}

We compute metrics that leverage the user goal and the dialog status at the end of the conversation to confirm that all necessary information has been exchanged, and the booked slots have the correct values. It is worth noticing that these metrics may be disrupted if one of the dialog components misinforms the other.\footnote{The dialog completion and success metrics are often criticized for being overly strict and incorporating heuristics that allow unfulfilled slots. Therefore, these metrics should be interpreted with caution~\cite{nekvinda-dusek-2021-shades}.}
\begin{itemize}
    \item \textbf{Succ Rate} evaluates matches for booked and informed slots. It is 1 if all the values match, 0 otherwise.
    \item \textbf{Compl Rate} is 1 if \textbf{all} the booked slots in the user goal have been filled (disregarding if the value is correct), 0 otherwise.
      \item \textbf{Book Rate} is the ratio of booked slots having a value matching the one in the user goal.
    \item \textbf{Inform Prec/Rec/F1} assess if the system provides the requested information accurately. True Positives (TP) are slots correctly mentioned, False Positives (FP) are slots with invalid values or only user-mentioned, and False Negatives (FN) are slots mentioned by the system with valid values but not in the goal.


    \item \textbf{DT} and \textbf{Successful DT} count the number of turns per dialog and per successful dialog respectively.  


\end{itemize}

\subsection{Lexical Diversity Metrics}


Generated user utterances can be formulaic and repetitive \cite{holtzman2020curious}. We evaluate the lexical diversity (LD) of the generated texts, using the training data as reference.
In particular, we use traditional metrics such as \textbf{user utterance (UUtt) length}, number of unique n-grams (\textbf{unigrams, bigrams, trigrams}) and selected metrics following \cite{nekvinda-dusek-2021-shades, duvsek2020evaluating}:
\begin{itemize}\setlength\itemsep{0em}
    \item Shannon Entropy \textbf{(SE)} measures information (in bits) in a random variable, i.e., an average number of bits needed to transmit the outcome of that variable, in our case, some token from all utterances \cite{manning1999foundations}.
\item Conditional bigram Entropy \textbf{(CE)} measures extra bits needed to describe the full bigram, given that we know the first word \cite{manning1999foundations}. Second word description needs more bits in diverse utterances.
\item Mean Segmental TTR (\textbf{MSTTR}) is the average value of TTR, the ratio of unique words (types) to total words (tokens), calculated on 50-words-long text segments~\cite{lu2012relationship}.
\item \textbf{MTLD} is derives from MSTTR to better handle shorter texts~\cite{mccarthy2010mtld}. 
\item \textbf{HDD} is a hypergeometric distribution function, which models, for each unique word in a text, the probability of encountering it in a random sample of 42 words drawn from that text \cite{wu1993accurate, mccarthy2010mtld}.
\end{itemize}

As a human LD baseline, we sample the training set 1,000 times to collect user utterances from 200 random dialogs, and report averaged metrics.

\subsection{Prompt Engineering}
Since LLM's outputs are extremely sensitive to the prompt's text, we explore several prompt design strategies: vary number of shots, supplement task description with persona/role modifiers, and test structured (bullet points) vs descriptive formats. Experiments and results are in Appendix~\ref{appendix:prompt_engineering}.

\subsection{Baselines and Models}

\begin{table*}[ht]

\caption{Goal fulfillment table. Best metrics values for each model are in bold and overall best results overall are marked with *. Runtime for GPT-3.5 depends on Azure OpenAI API load.} 
\centering\small
\begin{tabular}{
   lc
    rrrrrrrrc
}
\textbf{US Model} & \textbf{JS} & \begin{tabular}{p{0.6cm}} \textbf{Compl Rate} \end{tabular} & \begin{tabular}{p{0.6cm}} \textbf{Succ Rate} \end{tabular}& \begin{tabular}{p{0.6cm}} \textbf{Book Rate} \end{tabular} & \begin{tabular}{p{0.7cm}} \textbf{Inform Prec} \end{tabular} & \begin{tabular}{p{0.7cm}} \textbf{Inform Rec} \end{tabular}  & \begin{tabular}{p{0.7cm}} \textbf{Inform F1} \end{tabular} & \begin{tabular}{p{0.5cm}} \textbf{Succ DT} \end{tabular}  & \textbf{DT} & \begin{tabular}{l} \textbf{Avg Time}\\ \textbf{(minutes)} \end{tabular}\\
\toprule
ABUS-T & - & \textbf{*0.93} & \textbf{*0.83} &  \textbf{*0.85} & 0.84 & \textbf{*0.94} & 0.86 & 5.5 & \textbf{6.0} & 10\\
ABUS-D & - & 0.86 & 0.60 &  0.75 & \textbf{*0.87} & 0.90 & \textbf{*0.87} & \textbf{5.4} & 7.4 & 20\\
\midrule
GPT-3.5 & yes & \textbf{0.44} & 0.29 & 0.53 & 0.66 & 0.67 & 0.63 & \textbf{5.2} & \textbf{7.0} & 120\\
GPT-3.5  & no & 0.41 & \textbf{0.30} & \textbf{0.58} & \textbf{0.67} & \textbf{0.71} & \textbf{0.65} & 5.9 & 7.5 & \\\midrule
LLaMA  & yes & \textbf{0.05} & \textbf{0.07} & \textbf{0.25} & \textbf{0.34} & \textbf{0.25} & \textbf{0.27} & \textbf{4.6} & \textbf{6.9} & 80\\
LLaMA  & no & 0.04 & 0.04 & 0.23 & 0.27 & 0.21 & 0.21 & 12.0 & 8.1 & \\
\midrule
FLAN-T5  & yes & \textbf{0.14} & 0.13 & \textbf{0.32} & \textbf{0.43} & \textbf{0.35} & \textbf{0.36} & \textbf{*4.3} & \textbf{*5.3} & 60\\
FLAN-T5  & no & 0.12 & \textbf{0.15} & 0.28 & 0.41 & 0.34 & 0.34 & 4.4 & 5.6 & \\\bottomrule
\end{tabular}
\label{tab:goal_fulfillment}
\end{table*}

\paragraph{Dialog System}
As the dialog system to interact with the US, we use the best performing dialog system in {ConvLab-2}. It is a pipeline of a fine-tuned BERT~\cite{Devlin2019BERT} for intent and slot classification, rule-based DST and policy components, and template-based Natural Language Generation (NLG) component~\cite{zhu2020convlab2}.

\paragraph{User Simulator Baselines}
We use as baseline the agenda-based US~\cite{schatzmann2007agenda} from ConvLab-2.\footnote{We acknowledge the recent release of ConvLab-3~\cite{zhu2022convlab3}; however, it lacks support for evaluating the agenda-based user simulator, which is currently considered the strongest baseline for goal fulfillment metrics. Our attempts to utilize other US models~\cite{Lin2022GenTUSSU,Cheng2022finetuneUS} from ConvLab-3 did not yield comparable results as claimed by the authors. As a result, we decided to persist with ConvLab-2, leveraging its compatibility with the agenda-based US. ConvLab-2 has garnered significant usage and research attention, establishing itself as a prevalent benchmark for comparison and evaluation purposes.} We consider two variants: with a templated NLG and with data-driven NLG~\cite{wen2015nlg} for generating the user utterances. We refer to these user simulator models as \textbf{ABUS-T} and \textbf{ABUS-D} respectively. 

\paragraph{Prompt-based User Simulator Models.}
As base models for the prompt-based simulators, we consider a three different large language models:
\begin{itemize}\setlength\itemsep{0em}
    \item \textbf{GPT-3.5}: OpenAI GPT-3.5 Turbo\footnote{\url{https://learn.microsoft.com/en-us/azure/cognitive-services/openai/concepts/models}} (ChatGPT), version 0301; 
    \item \textbf{FLAN-T5}: instruction-finetuned version of T5 with 3B parameters~\cite{chung2022flant5};
    \item \textbf{LLaMa}: the auto-regressive model LLAMA with 7B parameters~\cite{Touvron2023LLaMAOA}.
    
\end{itemize}


As we track the user goal for evaluation purposes (Section~\ref{sec:architecture}), a prompt-based model uses a BERT-based NLU model to translate user and system utterances in a structured form. 


\subsection{Hyperparameters}
We explore different hyperparameters for the LLMs behind the prompt-based US. The prompt schema selected is presented in Table~\ref{tab:example_prompt}, utilizing 2 example shots. 
The reported results are based on experiments carried out with the best hyperparameters identified through the success rate (Succ Rate) metric. We set the temperature to 0.8 for LLaMa, 1 for GPT-3.5, and 0.9 for FLAN-T5. Each experiment consisted of 200 seeded dialogs, employing either Jaccard similarity (JS) or random sampling, and the results were averaged.

The experiments were run on a machine with an NVIDIA RTX GeForce A6000 GPU, CUDA v11.6, and 48GB RAM.

\section{Results}


\subsection{Goal Fulfillment Performance} \label{sec:exp_goal_fulfill}
\begin{table*}[ht]
\caption{Lexical diversity results. 
Best metrics values for each model are in bold. The best results overall are marked with *.}
\centering\small
\begin{tabular}{
    lrc
    rrrr
    rrrrrr
}
\textbf{US Model} & \textbf{T} & \textbf{JS} & \textbf{\#UUtt} & \begin{tabular}{p{0.5cm}} \textbf{UUtt Length} 
\end{tabular} &  \begin{tabular}{p{0.5cm}} \textbf{Uni-grams} 
\end{tabular} &  \begin{tabular}{p{0.5cm}} \textbf{Bi-grams} 
\end{tabular} & \begin{tabular}{p{0.5cm}} \textbf{Tri-grams} 
\end{tabular} & \textbf{SE} & \textbf{CE} &
 \begin{tabular}{p{0.6cm}} \textbf{MSTTR} 
\end{tabular} & \textbf{HDD} & \textbf{MTLD} \\
\toprule
Training set & -  & - & 1,345 & 12.6 & 919 & 4,505 & 7,678 & 7.3 & 3.0 & 0.76 & 0.78 & 61.0 \\
\midrule
ABUS-T &  - &  - & 1,191 & \textbf{*17.4} & \textbf{527} & \textbf{2,140} & 3,867 & \textbf{6.9} & \textbf{2.5} & \textbf{*0.72} & \textbf{0.75} & \textbf{46.9} \\
ABUS-D & - & - & 1,470 & 9.8 & 327 & 837 & 1,225 & 6.1 & 1.2 & 0.58 & 0.67 & 28.0 \\
\midrule
GPT-3.5 & 1.0 & yes & 1,395 & \textbf{16.7} & \textbf{876} & \textbf{3,934} & 6,970 & \textbf{7.3} & \textbf{2.8} & \textbf{0.70} & \textbf{0.78} & 47.4 \\
GPT-3.5 & 1.0 & no & 1,492 & 16.4 & 871 & 3,893 & \textbf{6,985} & \textbf{7.3} & \textbf{2.8} & \textbf{0.70} & \textbf{0.78} & \textbf{*48.3} \\
\midrule
LLaMA & 0.8 & yes & 1,381 & \textbf{11.0} & \textbf{860} & 3,537 & 5,599 & \textbf{7.1} & \textbf{2.7} & \textbf{0.61} & \textbf{*0.80} & \textbf{31.9} \\
LLaMA & 0.8 & no & 1,620 & 10.8 & 845 & \textbf{3,626} & \textbf{5,928} & \textbf{7.1} & \textbf{2.7} & 0.60 & \textbf{*0.80} & 30.6 \\
\midrule
FLAN-T5 & 0.7 & yes &  1,129 & 13.0 & 755 & 2,910 & 4,710 & 7.0 & 2.4 & 0.63 & 0.74 & 36.6 \\
FLAN-T5 & 0.7 & no & 1,086 & 13.6 & 732 & 2,995 & 4,962 & 7.0 & 2.5 & 0.64 & 0.74 & 37.6 \\
FLAN-T5 & 0.8 & yes & 1,159 & 13.9 & 897 & 3,733 & 6,030 & 7.2 & 2.7 & 0.66 & 0.76 & 40.0 \\
FLAN-T5 & 0.8 & no & 1,159 & \textbf{14.0} & 853 & 3,774 & 6,285 & 7.2 & 2.8 & 0.67 & 0.77 & 41.1 \\
FLAN-T5 & 0.9 & yes & 1,061 & 13.9 & \textbf{*994} & 4,370 & 6,977 & \textbf{*7.4} & \textbf{*3.0} & \textbf{0.70} & 0.77 & 44.8 \\
FLAN-T5 & 0.9 & no & 1,123 & 13.8 & 949 & \textbf{*4,440} & \textbf{*7,176} & \textbf{*7.4} & \textbf{*3.0} & \textbf{0.70} & \textbf{0.78} & \textbf{45.5}\\
\bottomrule
\end{tabular}
\label{tab:diversity}
\end{table*}


Let us note that the goal fulfillment metrics do not directly evaluate the user simulator, but instead quantify the effectiveness of the communication between the user simulator and the system, with the system's imperfect NLU component as a confounding factor. 

Table~\ref{tab:goal_fulfillment} presents the results, with the ABUS agenda-based models clearly outperforming the prompt-based models, as expected. ABUS-T's superior performance can be attributed to its perfect tailoring for system communication. On the other hand, the replacement of NLG in ABUS-D with a data-driven component results in a significant drop in the successful dialog rate.
Among the prompt-based models, GPT-3.5 demonstrates best performance, followed by FLAN-T5 and LLaMa. 
Regarding the Jaccard sampling (JS), it appears to be more beneficial for the smaller models LLaMa and FLAN-T5, potentially encouraging them to adhere more closely to the example dialogs.

In Section~\ref{sec:dialog_anticipator_results}, we delve into an analysis of the issues contributing to dialog breakdowns, which subsequently lead to low goal fulfillment scores. Yet, the system itself plays a significant role in the low success rate. Indeed, when the user's text significantly deviates from the NLU training data, the successful rate tends to decrease.





\subsection{Lexical Diversity Performance} \label{sec:exp_language_diversity}
Table~\ref{tab:diversity} presents the results of our LD analysis for the agenda and prompt-based models. We also investigate the effect of temperature~(\textbf{T}) by varying its values (0.7, 0.8, 0.9) with FLAN-T5.

LLMs generate utterances that match or even exceed the diversity of sampled training data (human utterances) and rule-based models. {GPT-3.5} produces more diverse utterances than ABUS-T in terms of SE, CE, HDD, and MTLD metrics, even surpassing the sampled training set in terms of SE and HDD. In contrast, ABUS-D has extremely low diversity scores, indicating a small vocabulary and limited language variability. The prompt-based models have significantly higher unique unigram, bigram, and trigram counts than ABUS-T and ABUS-D, approaching the counts of the training set. Thus, while LLMs have smaller goal fulfillment scores than agenda-based models, they employ richer vocabulary, which is especially useful when testing the NLU component. 
Among prompt-based models, GPT-3.5 generates the longest average user utterance, followed by FLAN-T5 and LLaMA, and is comparable to ABUS-T. In comparison, user utterances in the sampled training set conversations are approximately four tokens shorter than those of GPT-3.5 and ABUS-T, indicating that successful USs tend to produce longer utterances to increase the chances of achieving user goals.

We hypothesize that higher temperatures will result in a text with low-probability, more diverse words, thus, increasing LD metrics. Our results from FLAN-T5 confirm this strong trend. We also find that JS does not outperform random sampling in terms of LD, which is not surprising since we expect prompting the LLMs with random examples to increase the diversity of generated utterances.


\subsection{Error Analysis} 
\label{sec:dialog_anticipator_results}

Through a manual inspection of the results, we uncover several issues arising from both the user and system perspectives. This qualitative analysis corroborates the findings obtained from automatic measures. In particular, GPT-3.5 seems to have the most natural conversations, followed by FLAN-T5 and LLaMa. It is worth noting that the conversational interface focus of GPT-3.5 and the instruction-based fine-tuning of FLAN-T5 likely enable them to better adhere to the requirements compared to LLaMa, which solely functions as a language model. In the subsequent sections, we provide a detailed examination of some of these identified issues using examples. To maintain clarity and organization, we include the example conversations in Appendix~\ref{app:example_dialogs}.

\paragraph{System NLU misclassifications} As the simulator is used without any finetuning, the system NLU may have some difficulties in understanding what the user means to say. This is extremely useful because it allows us to detect misclassifications from the system NLU. Indeed, we notice that some of the user's statements do not correspond to any dialog act tuple. Examples include utterances such as ``That is all I needed. I am not interested in anything else'' or ``No, I do not have another destination in mind. Thank you for the help''. 

Although these misclassifications do not significantly impact the overall dialog, the system NLU struggles to understand other sentences, leading to dialog breakdowns~\cite{terragni-etal-2022-betold}. For example, the system NLU fails at understanding other sentences like ``How much is it?'', ``Yes, please. I need to reserve it now'', and ``May I have their telephone number, please?''. These misclassifications can be problematic because, when the system receives an empty dialog act list, it responds with a \textit{bye} intent, e.g., ``Okay, thank you. Have a good day.''. This behavior sometimes tricks a LLM-based simulator into \textit{believing} that the dialog has ended, and thus, it prematurely terminates the dialog before accomplishing their target goal.

\paragraph{User simulator fails to mention all the requirements} Another issue stems from the user simulator's failure to mention all the requirements. In contrast to an agenda-based simulator, a prompt-based US may easily give up after a dialog breakdown, deviating from the requirement to fulfill all the objectives described in the prompt. We illustrate this issue in Tables~\ref{tab:example_dialog_flant5_giveup} and \ref{tab:example_dialog_gpt_forgets}, where FLAN-T5 \textit{gives up} to continue the conversation and GPT-3.5 \textit{forgets} to ask for the price range, despite it being necessary according to the user goal. 
Although termination before goal completion occurs in less than 45\% of the conversations for prompt-based models (according to Table~\ref{tab:goal_fulfillment}), GPT-3.5 exhibits higher completion rates due to its more persistent behavior, as shown in Table~\ref{tab:example_dialog_gpt_insisting}.

\paragraph{User simulator generates hallucinations or degenerated text} 
We observed instances of hallucinations or degenerated text generated by the user simulator. Hallucinations refer to the production of text that deviates from the intended meaning~\cite{hallucinations}. 
For example, Table~\ref{tab:example_dialog_flant5_hallucination} demonstrates a case where the user arbitrarily expresses disinterest in Chinese food, despite it not being mentioned in their original goal. Additionally, we identified occurrences where the model repetitively produced the same utterance, as evident in Table~\ref{tab:example_dialog_llama_divergence}. In some cases, the user simulator even assumed the role of an assistant, disrupting the dialog. This behavior can be observed in Table~\ref{tab:example_dialog_flant5_assistant} and~\ref{tab:example_dialog_flant5_assistant_hallucinations}, in the latter example, the user also re-uses slot information from the previous prompt. The issue likely arises from the training data for LLMs, which typically emphasizes emulating assistant behavior rather than accurately representing user behavior.

\section{Discussion and Conclusions} 
In this study, we examined the effectiveness of an in-context learning user simulator for a task-oriented dialog system using LLMs. However, there are certain limitations to consider. Firstly, the lack of open-source availability of a specific LLM used, GPT-3.5, can hinder replicability of the experiments. Additionally, the opaque details of the pre-training and fine-tuning datasets make it difficult to assess whether these models have been exposed to examples from datasets like MultiWOZ~2.1 or very similar ones~\cite{bender2021dangers}. Although we evaluated our approach on a dataset that encompasses various domains and the few-shot learning capabilities of these models suggest that similar results can be obtained on different datasets, it is possible that the models may not generalize well to specific domains or low-resource languages. Moreover, LLMs often generate hallucinations~\cite{hallucinations}, and while prompt-based approaches offer flexibility, they can lead to unpredictable and potentially inappropriate responses. Care must be taken when deploying these models to avoid misleading outputs~\cite{Brown2020GPT3,bender2021dangers}.

Despite these challenges and limitations, our study demonstrated that in-context learning user simulation generates diverse language valuable for testing a dialog system. Rule-based models outperform prompt-based approaches in goal fulfillment metrics but produce stilted conversations. Prompt-based approaches enable exploration of unexpected paths and they just require the definition of a user goal and a few example conversations. While our models did not achieve the same success as the agenda-based model, our qualitative analysis identified critical issues in system robustness, including user simulator and dialog system errors. These errors include NLU misclassifications from the dialog system side, while, from the user side, we noticed hallucinations, repetitions and lack of consistency with the user goal. Addressing the discussed limitations and further research can advance the field towards reliable and effective user simulators across diverse domains and languages.

\bibliographystyle{acl_natbib}
\bibliography{biblio}

\appendix

\appendix

\section{Prompt Engineering}\label{appendix:prompt_engineering}
One of the advantages of LLM-powered US is that there is no need for fine-tuning. However, it is crucial to craft an adequate prompt in natural language which can be understood by the LLM~\cite{Brown2020GPT3,Schick2020ItsNJ,Schick2020ExploitingCF}. 
In this work, the prompt consists of a task description which grounds the LLM with a persona and certain roles to fulfill, and a number of dialog examples (shots) with a pre-pended requirements sentence derived from the user goals.

\subsection{Experimental Setup} \label{appendix:prompt_engineering_methods}
We explore different prompts in order to optimize the fulfillment of the user goal. 
In particular, we investigate different alternatives for the general task description and the structure of the user goal requirements. As before, we run our experiments for 200 dialogs and average the results. We focus this analysis only on FLAN-T5.  Instead of performing a complete grid search, we fix the values of all the parameters except for one parameter, for each dimension that we want to investigate. We detail the different dimensions in the following sections. Let us also notice that these experiments were run in parallel to other experiments. Back then, the temperature was set to 0.8. Only after, we found out that 0.9 temperature resulted in better performance. For this reason, the success rate and lexical diversity for 2 prompts with the default template do not coincide with the ones reported in Tables~\ref{tab:diversity} and ~\ref{tab:goal_fulfillment}.    
\begin{table*}[ht]
\centering\small
\begin{tabular}{cp{0.7in}|ccccc}
\# Shots & Bullet Point Requirements &  \multicolumn{5}{c}{Task Descriptions} \\
\toprule
& & default & default + domains & extra personality & minimal & none \\
\midrule

0 & FALSE & 0.08 (0.68) & 0.07 (0.68) & {0.15} (0.65) & 0.01 (0.71) & 0.07 (0.64) \\
1 & FALSE & 0.10 (0.67) & 0.10 (0.66) & 0.12 (0.62) & 0.05, (0.70) & 0.06 (0.65) \\
 & TRUE & 0.08 (0.69) &  &  &  &  \\
2 & FALSE & 0.12 (0.66) & 0.10 (0.67) & 0.11 (0.62) & 0.10 (0.69) & 0.07 (0.63) \\
 & TRUE & 0.11 (0.68) &  &  &  &  \\
3 & FALSE & {0.14} (0.67) & {0.15} (0.66) & 0.13 (0.61) & 0.07 (0.69) & 0.09 (0.66) \\
 & TRUE & 0.12 (0.68) &  &  &  &  \\

\midrule
4 & FALSE & 0.13 (0.66) &  &  &  & 0.09 (0.65) \\
5 & FALSE & 0.12 (0.66) &  &  &  &  {0.10} (0.66) \\
6 & FALSE & 0.11 (0.65) &  &  &  & 0.08 (0.65)\\ \bottomrule
\end{tabular}
\caption{Prompt engineering results after running 200 identical conversations under varying number of shots/examples, 5 task description flavors, requirements representation (structured bullet points vs descriptive). Reported scores are dialog success ratio and MSTTR (in parenthesis). In 0-shot setting, default + domains prompt is equivalent to default, because domains are derived from the shots' goals. US is modeled by FLAN-T5.}
\label{tab:prompt}
\end{table*}

\subsubsection{Task description}

We want the LLM to enact a \textbf{role} of the customer, with a certain \textbf{persona} attributes that make them more likely to successfully book appointments.
Examples of persona modifiers include: \textit{"You are a picky tourist"}, \textit{"try to rephrase your request until you are sure the ASSISTANT understood you"}. Representative role modifiers: \textit{"Complete the conversation as a CUSTOMER"}, \textit{"Be precise with the REQUIREMENTS, clear and concise"}.

Next, in order to force the LLM to generate text for the subset of domains which appear in user goal requirements (out of 5 possible domains), \textbf{domain names} were extracted from the selected example shots, and included inside the default task description (see default + domains), 

Here are all considered task descriptions: 
\begin{enumerate}
    \item \textbf{Default:} \textit{These are examples of conversations between a CUSTOMER and an ASSISTANT. Complete the conversation as a CUSTOMER. The conversation is about making a reservation or asking for information according to the REQUIREMENTS. You must fulfill all REQUIREMENTS in the given order. Be precise with the REQUIREMENTS, clear and concise.} 
    \item \textbf{Default + domains:} \textit{These are examples of conversations between a CUSTOMER and an ASSISTANT. Complete the conversation as a CUSTOMER. The conversation is about making a reservation or asking for information about \texttt{<domain names>} according to the REQUIREMENTS. You must fulfill all REQUIREMENTS in the given order. Be precise with the REQUIREMENTS, clear and concise.} 
    \item \textbf{Default + extra personality:} \textit{These are examples of conversations between a CUSTOMER and an ASSISTANT. You are a picky tourist and talking to an ASSISTANT at a tourist office. Complete the conversation as a CUSTOMER. The conversation is about making a reservation or asking for information according to the REQUIREMENTS. Make sure you mention all REQUIREMENTS. You must satisfy the REQUIREMENTS in the given order. The tourist office ASSISTANT may have difficulties understanding you and may respond out of context. If the ASSISTANT responds with an unexpected answer try to rephrase your request until you are sure the ASSISTANT understood you and gave you all the correct answers.} 
    \item \textbf{Minimal:} \textit{These are examples of conversations between a CUSTOMER and an ASSISTANT. Complete the conversation as a CUSTOMER.} 
    \item \textbf{None:} \textit{<no task description>} 
\end{enumerate}

\subsubsection{Structured vs descriptive representation of Requirements}

We test two formats of representing the goal's requirements: structured and descriptive. The former representation uses new-line separated bullet points with each requirement, and the latter simply lists the requirements in a series of sentences, as generated by the pre-defined templates in ConvLab2.

\subsubsection{Number of example shots}
We tested the effect of adding from 0 to 6 example dialogs into the prompt.

\subsection{Prompt Engineering Results} \label{appendix:prompt_engineering_results}

Results from experiments with FLAN-T5 model are shown in Table \ref{tab:prompt}. Reported scores are slot dialog success ratio and, in parenthesis, MSTTR.

Our experiments show that adding domain names to the task description does not improve performance, while enhancing the persona and role description with extra instructions leads to significant improvements, even in a 0-shot setting. The minimal task description performs worse than the default but generates the most lexically diverse user utterances. Using no task description and relying on example shots consistently outperforms the minimal description setting, but this result would require additional manual inspection.

We also found that increasing the number of shots improves performance up to a certain point, after which the positive effect plateaus. We suspect that longer prompts with multiple example dialogs may exceed the LLM's input space, leading to decreased performance. Representing the goal's requirements with structured bullet points did not significantly improve task fulfillment metrics.

Overall, we can observe that the user simulator is highly sensitive to the prompt and even small changes in the prompt can lead to changes in the results.

\section{Examples of Generated Dialogs}\label{app:example_dialogs}
We include the example conversations in this Section to mantain clarity and organization. 




\begin{table*}[h]
\caption{FLAN-T5 dialog. The user simulator \textit{gives up} and ends the dialog.}
\resizebox{1\linewidth}{!}{%
\begin{tabular}{p{18cm}}
\toprule
REQUIREMENTS: You are looking for a place to stay. The hotel should be in the east. The hotel should have a star of 4. The hotel should include free parking. The hotel should include free wifi. Once you find a hotel, make sure you get phone number. You are looking forward to trying local restaurants. You are looking for a particular restaurant. Its name is called taj tandoori. Make sure you get postcode\\
\\
USER: I am looking for a hotel on the West side. I have a budget of 100 dollars per night. Can you suggest a hotel with 3 stars or Above ?\\
ASSISTANT: How about hobsons house ? It is moderate . They have free parking . It is in the west part of town . It does have 3 stars . It is a guesthouse . The wifi is included .\\
USER: Yes please. I would like to book a double room. It is a guesthouse . The price is 100 dollars per night. I would like to book it now.\\
ASSISTANT: Hobsons house is available would you like to try that ? The parking is free . They have a 3 Star rating. It is in the west part of town . It has wifi . It is a guesthouse . Its listed as moderate .\\
USER: Yes please . I would like to book a double room. Do they have free parking ?\\
ASSISTANT: What about hobsons house? It does include free parking.\\
USER: That sounds good . They have parking . The price is 100 dollars per night. I would like to book now.\\
ASSISTANT: Sorry there are none in the west . It does n't need to have parking . I am sorry , there are not 3 stars available . I ca n't find any guesthouse that fit your criteria , i ' m sorry .\\
USER: Ok . That is all correct . I will end my search here . Thank you.\\
\bottomrule                                         
\end{tabular}}
\label{tab:example_dialog_flant5_giveup}
\end{table*}

\begin{table*}[h]
\caption{GPT-3.5 dialog. The user simulator does not ask for the price range of the restaurant.}
\resizebox{1\linewidth}{!}{%
\begin{tabular}{p{18cm}}
\toprule
REQUIREMENTS: You are looking forward to trying local restaurants. The restaurant should be in the expensive price range. The restaurant should be in the centre. Once you find the restaurant you want to book a table for 2 people at 11:30 on sunday
\\
\\
USER: Hi, I'm interested in trying some local restaurants in the city centre.
\\ASSISTANT: There are 69 different places that match your description . I would suggest mahal of cambridge . It is in the centre area .\\
USER: Sounds good. Can I make a reservation for 2 people at 11:30 on Sunday ?\\
ASSISTANT: Your reference number is 00000000 .\\
USER: I'm sorry, can you please repeat the reference number? I didn't catch that.\\
ASSISTANT: Your reference number is 00000000 .\\
USER: Thank you. That's all for now. Have a good day!\\
\bottomrule                                         
\end{tabular}}
\label{tab:example_dialog_gpt_forgets}
\end{table*}

\begin{table*}[h]
\caption{GPT-3.5 dialog. The user simulator is \textit{insistent} to fulfill their goals.}
\resizebox{1\linewidth}{!}{%
\begin{tabular}{p{18cm}}
\toprule
USER: Wait, how do I make a reservation at the city centre north b and b?\\
ASSISTANT: It offers free parking. \\
USER: Yes, I know. But how do I book a room at the city centre north b and b?\\
ASSISTANT: How about city centre north b and b ?\\
USER: Yes, that's the one I want to book. How do I make a reservation? \\
ASSISTANT: You are more than welcome!\\
USER: Um, excuse me, how can I make a reservation at the city centre north b and b? \\
ASSISTANT: What about city centre north b and b? The wifi is included. It does include free parking. It is a guesthouse.  \\
USER: Yes, I want to make a reservation at the city centre north b and b. How do I do that?    \\
ASSISTANT: What about city centre north b and b? It does have 0 stars. The parking is free. It is indeed in the north. The price range is cheap. It is a guesthouse.               \\
USER: I understand that, but how do I make a reservation for a room at city centre north b and b?                                                                                   \\
ASSISTANT: How about city centre north b and b ? It does include free parking. It is a guesthouse. They have a 0 Star rating. They are located in the north. This is an cheap hotel. \\
USER: Yes, that's the one I want to book. Can you please make a reservation for 8 people starting from Thursday for 4 nights? \\
ASSISTANT: Reference number is: 00000013. \\
USER: Thank you! Now, I also need to book a taxi from the hotel to the restaurant. We need to leave the hotel after 9:30pm. \\
... \textit{conversation continues}\\
\bottomrule
\end{tabular}}
\label{tab:example_dialog_gpt_insisting}
\end{table*}

\begin{table*}[h]
\caption{LLaMA dialog. The user simulator keeps repeating the same text.}
\resizebox{1\linewidth}{!}{%
\begin{tabular}{p{18cm}}
\toprule
REQUIREMENTS: You are looking forward to trying local restaurants. The restaurant should serve italian food. The restaurant should be in the centre. Once you find a restaurant, make sure you get phone number, address. You are looking for a place to stay. The hotel should be in the east. The hotel should include free wifi. The hotel should be in the type of hotel. Once you find the hotel you want to book it for 4 people on tuesday for 3 nights\\
\\
CUSTOMER: I am looking forward to try local restaurants . The restaurant should be in the centre . The restaurant should serve Italian food.\\
ASSISTANT: There are 9 available restaurants . Zizzi cambridge matches your description . Would you like italian food ?
\\CUSTOMER: That sounds good. Can you tell me about the restaurant?
\\ASSISTANT: There are 9 restaurants in that area that fit that criteria. Caffe uno sounds like it might be what you are looking for . It is located in the centre .
\\CUSTOMER: That sounds good. What is the postcode for the restaurant?
\\ASSISTANT: Their postcode is cb21ab.
\\CUSTOMER: Is the restaurant open on tuesday ?
\\ASSISTANT: Reference number is : 00000000 .
\\CUSTOMER: Can you tell me about the restaurant .
\\ASSISTANT: There are 9 available restaurants . I would suggest zizzi cambridge .
\\CUSTOMER: Can you tell me about the restaurant ?
\\ASSISTANT: There are 9 restaurants in that area that fit that criteria. Pizza hut city centre has some great reviews .
\\CUSTOMER: Can you tell me about the restaurant .
\\ASSISTANT: There are 9 restaurants in that area that fit that criteria. Pizza express Fen Ditton has some great reviews . The address is Jesus Lane Fen Ditton .
\\CUSTOMER: Can you tell me about the restaurant ?
\\ASSISTANT: I have 9 different restaurants I can give you some information for . They are all pretty good . I have a place called pizza express Fen Ditton , does that sound like something you would enjoy? The reference number is 00000075 .
\\CUSTOMER: Can you tell me about the restaurant ?
\\
... \textit{conversation continues}
\\
\bottomrule
\end{tabular}}
\label{tab:example_dialog_llama_divergence}
\end{table*}

  \begin{table*}[h]
\caption{FLAN-T5 dialog. The user simulator starts acting as the assistant.}
\resizebox{1\linewidth}{!}{%
\begin{tabular}{p{18cm}}
\toprule
REQUIREMENTS: You are looking forward to trying local restaurants. The restaurant should be in the centre. The restaurant should serve chinese food. The restaurant should be in the expensive price range. Once you find a restaurant, make sure you get phone number. Once you find the restaurant you want to book a table for 2 people at 18:45 on sunday
\\
\\
CUSTOMER: I am looking forward to trying local restaurants . I am looking for a Chinese restaurant in the centre that is in the expensive price range .
\\ASSISTANT: There are 4 available restaurants . I would suggest ugly duckling .
\\CUSTOMER: How much is it ?
\\ASSISTANT: Thank you for contacting the help desk . Have a great day .
\\CUSTOMER: It is 57 pounds per person and they are taking reservations from 18:45 .
\\ASSISTANT: Booking was successful . Reference number is : 00000038 .\\
\bottomrule                                         
\end{tabular}}
\label{tab:example_dialog_flant5_assistant}
\end{table*}

\begin{table*}[h]
\caption{FLAN-T5 dialog. The user simulator starts acting as the assistant and uses information from the example prompt. Example 3 corresponds to the generated conversation.}
\resizebox{1\linewidth}{!}{%
\begin{tabular}{p{18cm}}
\toprule
...
\\
Example 2:
\\REQUIREMENTS: You were robbed and are looking for help. Make sure you get postcode and address
\\
\\CUSTOMER: Please help me ! I 've just been robbed !
\\ASSISTANT: The telephone number for the Parkside Police Station is 01223358966 .
\\CUSTOMER: Thank you , could you also provide me the postcode and address ?
\\ASSISTANT: They are in Parkside , Cambridge CB11JG .
\\CUSTOMER: Thank you so much , I appreciate your help !
\\ASSISTANT: Is there anything else I can help you with ?
\\CUSTOMER: No that is it . Goodbye .
\\ASSISTANT: Goodbye . I hope the police are able to help you , I 'm sorry this happened to you .
\\\\
Example3:
\\REQUIREMENTS: You were robbed and are looking for help. Make sure you get postcode, address
\\
\\CUSTOMER: I was in a car accident dispute and am looking for help .
\\ASSISTANT: Thank you for contacting us and have a nice day .
\\CUSTOMER: Thanks for contacting us . The phone number for the police is 01223358966 .
\\ASSISTANT: Thank you for contacting us and have a nice day .
\\CUSTOMER: The telephone number for the Parkside Police Station is 01223358966 .
\\ASSISTANT: Ok , thank you . Have a good day .
\\CUSTOMER: Thank you for contacting us . I was in a car accident dispute and am looking for help .
\\ASSISTANT: You are more than welcome !.\\
\bottomrule                                         
\end{tabular}}
\label{tab:example_dialog_flant5_assistant_hallucinations}
\end{table*}

\begin{table*}[h]
\caption{FLAN-T5 dialog. The user simulator generates hallucinations by conflicting with their goal.}
\resizebox{1\linewidth}{!}{%
\begin{tabular}{p{18cm}}
\toprule
REQUIREMENTS: You are looking forward to trying local restaurants. The restaurant should serve lebanese food. The restaurant should be in the north. The restaurant should be in the expensive price range. If there is no such restaurant, how about one that serves chinese food. Once you find a restaurant, make sure you get postcode
\\
\\USER: I would like to find a nice restaurant in the north that has Lebanese food and is expensive .
\\ASSISTANT: I ' m sorry I have no restaurants serving lebanese food. I do n't have anything in the expensive range that fits that criteria . There are none in north of town .
\\USER: Well I 'm not interested in a chinese restaurant because I 'm looking forward to trying local restaurants .
\\ASSISTANT: I have nothing with lebanese . Do you have another preference ? There are no expensive ones . I have n't found any in the north.
\\USER: I 'm not interested in anything . I 'm not interested in any chinese restaurants . I also want to get the address and postcode of the restaurants .
\\ASSISTANT: I am unable to find any lebanese restaurants in town . No expensive restaurant. I did not find any restaurants in north.
\\USER: I think I would just like to go to a local restaurant that serves Lebanese food . Have you found anything?
\\ASSISTANT: I am unable to find any lebanese restaurants in town . There are no expensive ones . I did not find any restaurants in north.
\\USER: What 's up ? I want a local restaurant that serves Lebanese food and is expensive . I guess I will try elsewhere.
\\ASSISTANT: I did not find any lebanese restaurants . No expensive restaurant. There no such restraunts in north. I ' m sorry I have no restaurants serving lebanese food . No expensive restaurant. I am sorry but there are no restaurants that fit that criteria in the north .
\\USER: I need the address and postcode of the restaurants.
\\ASSISTANT: There are no lebanese food places , shall I run another search ? I do n't have anything in the expensive range that fits that criteria . I have n't found any in the north.
\\USER: Well I need the address and postcode of the restaurants
\\ASSISTANT: I have nothing with lebanese . Do you have another preference ? I do n't have anything in the expensive range that fits that criteria . I did not find any restaurants in north.
\\USER: What 's up ? I want to find a local restaurant that serves lebanese food and is expensive . Have you found anything?
\\ASSISTANT: I do not have anything in that price range for lebanese . Another criteria perhaps ? No expensive restaurant. I am sorry but there are no restaurants that fit that criteria in the north .
\\USER: Ok well there is nothing there , oh and please get me the address and postcode
\\
... \textit{conversation continues}
\\\bottomrule
\end{tabular}}
\label{tab:example_dialog_flant5_hallucination}
\end{table*}

\end{document}